\documentclass[review]{elsarticle}
\usepackage{amsmath}
\usepackage{amsthm}
\usepackage{amssymb}
\usepackage{lineno}

\usepackage{adjustbox}
\usepackage{bm}
\usepackage{balance}
\usepackage{dblfloatfix}
\usepackage[vlined,linesnumbered,titlenumbered,ruled]{algorithm2e}
\usepackage{array, boldline, makecell, booktabs} 
\usepackage{tabularx}{\tiny}
\usepackage{color, colortbl}
\usepackage{graphicx}
\usepackage{epsfig}
\usepackage{mathrsfs}
\usepackage{multirow}
\usepackage[skip=2pt]{caption}
\usepackage{enumitem}
\usepackage{url}
\usepackage{subcaption}
\usepackage{apalike}
\usepackage{hyperref}

\newtheorem{definition}{Definition}

\newcommand{\mat}[1]{\boldsymbol{\mathrm{#1}}}

\newcommand{\bbR}{\mathbb{R}}

\newcommand*{\Dmodel}{D_{\mathrm{model}}}

\modulolinenumbers[5]

\journal{Neural Networks}

\begin{document}
\begin{frontmatter}

  \title{GeoMAE: Masking Representation Learning for Spatio-Temporal Graph Forecasting with Missing Values}

  \author[1,2,3]{Songyu~Ke\fnref{songyu}}
  \ead{songyuke@fzu.edu.cn}
  \author[4]{Chenyu~Wu}
  \ead{chenywu@swjtu.edu.cn}
  \author[5]{Yuxuan~Liang}
  \ead{yuxliang@outlook.com}
  \author[6]{Huiling Qin}
  \ead{qinhuiling@bnu.edu.cn}
  \author[2,3,4]{Junbo~Zhang\corref{ca}}
  \ead{msjunbozhang@outlook.com}
  \author[2,3,4]{Yu~Zheng}
  \ead{msyuzheng@outlook.com}

  \fntext[songyu]{This research was done when the first author was an intern at JD Intelligent Cities Research \& JD iCity under the supervision of the fifth author.}
  \cortext[ca]{Junbo Zhang is the corresponding author.}

  \affiliation[1]{
    organization={College of Computer and Data Science, Fuzhou University},
    city={Fuzhou},
    province={Fujian},
    country={China}
  }

  \affiliation[2]{
    organization={JD iCity, JD Technology},
    city={Beijing},
    country={China}
  }

  \affiliation[3]{
    organization={JD Intelligent Cities Research},
    city={Beijing},
    country={China}
  }

  \affiliation[4]{
    organization={School of Computing and Artificial Intelligence, Southwest Jiaotong University},
    city={Chengdu},
    province={Sichuan},
    country={China}
  }

  \affiliation[5]{
    organization={Hong Kong University of Science and Technology (Guangzhou)},
    city={Guangzhou},
    province={Guangdong},
    country={China}
  }
  \affiliation[6]{
    organization={Beijing Normal University},
    city={Zhuhai},
    province={Guangdong},
    country={China}
  }
  \begin{abstract}
    The ubiquity of missing data in urban intelligence systems, attributable to adverse environmental conditions and equipment failures, poses a significant challenge to the efficacy of downstream applications, notably in the realms of traffic forecasting and energy consumption prediction.
    Therefore, it is imperative to develop a robust spatio-temporal learning methodology capable of extracting meaningful insights from incomplete datasets. Despite the existence of methodologies for spatio-temporal graph forecasting in the presence of missing values, unresolved issues persist.

    Primarily, the majority of extant research is predicated on time-series analysis, thereby neglecting the dynamic spatial correlations inherent in sensor networks.
    Additionally, the complexity of missing data patterns compounds the intricacy of the problem.
    Furthermore, the variability in maintenance conditions results in a significant fluctuation in the ratio and pattern of missing values, thereby challenging the generalizability of predictive models.

    In response to these challenges, this study introduces GeoMAE, a self-supervised spatio-temporal representation learning model.
    The model is comprised of three principal components: an input preprocessing module, an attention-based spatio-temporal forecasting network (STAFN), and an auxiliary learning task, which draws inspiration from Masking AutoEncoders to enhance the robustness of spatio-temporal representation learning.

    Empirical evaluations on real-world datasets demonstrate that GeoMAE significantly outperforms existing benchmarks, achieving up to 13.20\% relative improvement over the best baseline models.
  \end{abstract}


\end{frontmatter}

\section{Introduction}
Spatio-temporal representation learning has emerged as a pivotal research area, underpinning various intelligent applications in smart cities that play crucial roles across multiple domains.
For instance, precise weather forecasting can significantly mitigate the detrimental impacts of natural disasters through early prevention; advanced traffic prediction systems help optimize traffic flow and substantially reduce congestion; environmental monitoring enables rapid identification of pollution hotspots within urban environments.

However, the prevalence of missing values in spatio-temporal data, attributable to sensor failures, network interruptions, human errors, and other issues during data acquisition, transmission, and storage, significantly complicates the representation learning process.
These missing values profoundly impact the complex temporal and spatial dependencies inherent in spatio-temporal data, thereby impeding the ability to discern underlying patterns.
Moreover, imputation of missing values, while often necessary for deep learning models, incurs additional computational costs and may introduce biases that adversely affect the models' generalization capabilities.
Consequently, effectively addressing missing values and learning representations from incomplete spatio-temporal data has become a critical research direction in this domain.

Current representation learning approaches for incomplete data can be broadly categorized into two principal paradigms.
The first approach treats missing value imputation as a distinct task, employing independent methods or models to accurately impute missing values before applying dedicated representation learning models for downstream tasks such as prediction or classification.
Traditional machine learning methods, including linear regression, K-Nearest Neighbors, random forests, clustering, and principal component analysis \cite{Haworth2012,Stekhoven2012,Ku2016clustering,Henrickson2015loopimputation}, alongside advanced deep learning techniques such as Variational Autoencoders (VAE) \cite{Fortuin2020GP-VAE,Nazabal2020VAE-Impuataion}, Generative Adversarial Networks (GAN) \cite{Yoon2018GAIN,Li2019MisGAN}, and Denoising Diffusion Probabilistic Models (DDPM) \cite{Tashiro2021CSDI}, have been deployed for data imputation.
This paradigm offers flexibility in selecting appropriate imputation techniques based on data characteristics and missing patterns.
However, it necessitates training both imputation and downstream task models, increasing computational complexity, while errors from the imputation model may propagate to downstream tasks, potentially degrading performance.

The alternative paradigm involves directly modeling incomplete data for downstream tasks, streamlining the processing pipeline and mitigating the impact of imputation errors.
For instance, decision tree-based models typically treat missing values as special categories requiring no additional handling.
Models based on Neural Ordinary Differential Equations (Neural ODEs) view irregularly spaced observations as multiple measurements of a continuous process, attempting to fit the data using Neural ODEs \cite{Chen2018NeuralODE,habiba2020neurodesinfomissing,Chang2023TN-ODE}.
Nevertheless, significant temporal irregularities in spatio-temporal data collection can adversely affect Neural ODEs, and their substantial computational demands challenge application to large-scale datasets.
The TriD-MAE model employs multi-period Temporal Convolutional Networks (TCNs) to capture temporal correlations in incomplete data, coupled with a masked autoencoding auxiliary task to enhance representation robustness for multivariate time series prediction \cite{Zhang2023TriD-MAE}.
The GinAR model introduces an imputation-aware attention mechanism to learn spatio-temporal dependencies from incomplete data for subsequent forecasting tasks \cite{Yu2024GinAR}.

\begin{figure}
  \centering
  \includegraphics[width=0.9\linewidth]{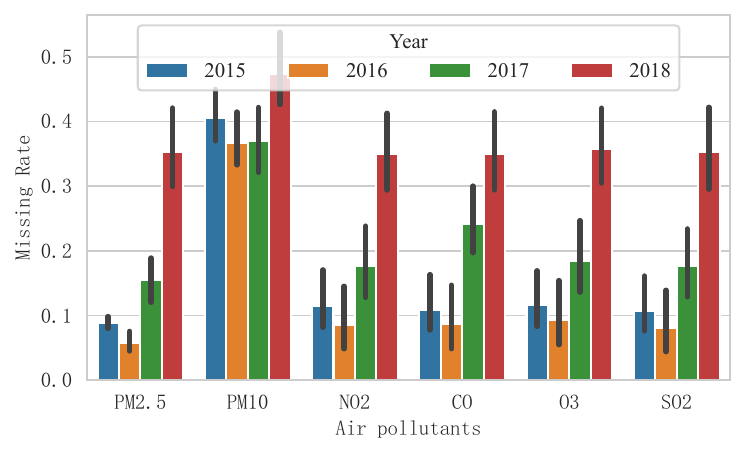}
  \caption{The changes on missing proportions of features (best viewed in the colored version)}
  \label{fig:variable-mising_prop}
\end{figure}

Despite these advances, constructing effective representation learning models for incomplete spatio-temporal data still faces significant challenges:

\begin{enumerate}
  \item \textbf{Complex and Dynamic Spatio-Temporal Dependencies}

        Spatio-temporal data exhibits intricate and dynamically evolving dependencies. Taking air quality prediction as an example, readings from a monitoring station correlate not only with its historical data but also with concurrent measurements from nearby stations. These spatial correlations are highly dynamic, influenced by geographic location, temporal factors, and external conditions such as wind direction and speed. Additionally, strong inter-variable correlations exist—for instance, sulfur dioxide and nitrogen oxides can transform into PM2.5 particles under specific conditions. Effectively modeling these temporal, spatial, and variable correlations amidst missing values remains a fundamental challenge for representation learning models.

  \item \textbf{Diverse Missing Patterns}

        Spatio-temporal data manifests various missing patterns, broadly classified into random missing, row missing, column missing, and block missing. Random missing, often caused by sporadic device errors or packet loss, typically affects individual time points and locations. Row missing occurs when all sensor readings are absent for a specific period, often due to system-wide failures. Column missing refers to continuous absence of one or more variables from a specific location, usually resulting from sensor maintenance or environmental damage. Block missing denotes simultaneous column missing across multiple sensors over extended periods, causing substantial information loss. Real-world datasets often present a mixture of these patterns, posing considerable challenges for representation learning.

  \item \textbf{Fluctuating Missing Conditions}

        In practical systems, data missingness closely relates to maintenance quality, which is directly constrained by operational budgets. During well-funded periods, systems are properly maintained, sensors operate reliably, and missing rates remain low. Conversely, budget constraints lead to inadequate maintenance, increased equipment failures, and consequently, degraded data quality with higher missing rates and more complex patterns. As illustrated in Figure~\ref{fig:variable-mising_prop}, which shows missing proportions for Beijing's air quality data across years, the missing rate in 2016 is notably lower than in other years, demonstrating this budget-driven phenomenon. Current research often overlooks these realistic fluctuations, typically assuming fixed missing patterns and rates, potentially compromising model generalization in real-world applications.
\end{enumerate}

To address these challenges and enhance spatio-temporal representation learning with incomplete data, we propose GeoMAE, a novel model for spatio-temporal graph data prediction with missing values. Our approach comprises three key components:

\begin{enumerate}
  \item \textbf{Input Preprocessing Module:} Employing random imputation strategies and specially designed masking matrices to mitigate distributional shifts caused by varying missing rates, enhancing model adaptability to different levels of data incompleteness.

  \item \textbf{Attention-based Spatio-Temporal Representation Network:} Leveraging spatio-temporal and spatial attention modules to efficiently capture complex dependencies and learn effective representations from incomplete data.

  \item \textbf{Multi-task Optimization with Attribute Masking:} Enhancing representation robustness by minimizing distances between representation vectors of the same spatio-temporal point under different masking patterns, reducing the impact of missing data noise.
\end{enumerate}

The main contributions of this work are four-fold:

\begin{enumerate}
  \item We present one of the first comprehensive investigations into the impact of varying missing data conditions (including rates and patterns) on representation learning performance, with targeted architectural designs to mitigate these effects.

  \item We propose a novel input processing module that effectively alleviates distributional shifts caused by varying missing rates, significantly improving model generalization capability.

  \item We design a self-supervised multi-task learning framework that constructs augmented samples through simulated missing scenarios, enhancing representation robustness against data incompleteness.

  \item Extensive experiments on real-world datasets demonstrate the superiority of our approach compared to state-of-the-art methods.
\end{enumerate}

\section{Related Works}
\subsection{Missing Value Imputation}

Spatio-temporal data often contains a significant amount of missing values due to various errors occurring during data acquisition, processing, transmission, and storage.
A range of methods have been developed for missing value imputation.
The simplest approach is mean imputation, which fills missing entries with the average value of the available data.
This method is straightforward and preserves the mean of the dataset. For data that is Missing Completely at Random (MCAR), mean imputation can maintain the original data distribution reasonably well.
Alternatively, using the most recent non-missing value can be a suitable choice for time-ordered data.
Additionally, various statistical methods are commonly employed, including least squares, naive Bayes, and principal component analysis \cite{Henrickson2015loopimputation, Tan2013, Tang2018}.

With advances in machine learning, algorithms such as K-Nearest Neighbors (KNN), Random Forests, and Generative Adversarial Networks (GANs) have been widely applied.
Missforest \cite{Stekhoven2012}, an R package, provides a general-purpose imputation method that first uses mean or mode for initial imputation, then iteratively trains random forest models on each feature with missing values to predict and update the imputations until convergence.
\citeauthor{Zhong2004Estimation} developed a genetic algorithm optimized with an artificial neural network for missing value estimation \cite{Zhong2004Estimation}.
\citeauthor{Zhao2015BayesianCPF} proposed a Bayesian Canonical Polyadic Decomposition (CPD) framework for incomplete tensor decomposition and recovery \cite{Zhao2015BayesianCPF}.
GAIN \cite{Yoon2018GAIN} adapted GANs for missing data imputation by generating imputations that follow the distribution of the observed data.

However, spatio-temporal data possesses unique temporal and spatial attributes, and often contains a high proportion of Not Missing at Random (NMAR) values, making the general methods mentioned above less suitable.
For example, mean imputation ignores the temporal proximity and periodicity inherent in spatio-temporal data \cite{zhang2017deep}.
Extreme weather events can cause sensor failures leading to NMAR missingness, and ignoring dynamic external factors like meteorological conditions can significantly reduce imputation precision.

Therefore, researchers have designed models specifically tailored for spatio-temporal data.
\citeauthor{Haworth2012} utilized a KNN-based method to recover missing traffic data by leveraging information from neighboring road segments \cite{Haworth2012}.
\citeauthor{Ku2016clustering} introduced a clustering-based imputation method that uses Stacked Denoising Autoencoders to obtain high-dimensional representations of traffic data, followed by clustering to identify similar samples for imputation \cite{Ku2016clustering}.
\citeauthor{Yi2016ST-MVL} proposed the ST-MVL model, which captures spatio-temporal characteristics from four aspects: spatial proximity, temporal proximity, spatial similarity, and temporal similarity, combining them using a linear model to generate final imputations \cite{Yi2016ST-MVL}.
GANs have also been adapted for spatio-temporal data generation and imputation.
\citeauthor{Gao2022STGAN-Survey} surveyed various GAN architectures suitable for time series and spatio-temporal data \cite{Gao2022STGAN-Survey}.
To mitigate GANs' training instability, \citeauthor{Qin2021Network} introduced the ST-SCL model, which incorporates a regression loss alongside the adversarial loss, simplifying training and improving performance on traffic data imputation tasks \cite{Qin2021Network}.

\subsection{Directly Modeling Incomplete Data}

The aforementioned research primarily treats missing value imputation as a separate machine learning task.
Subsequent prediction, classification, or recommendation tasks require fitting new models on the imputed data.
While this two-step approach is modular and widely applicable, it may introduce bias or noise, and imputation errors can propagate to downstream tasks.
Therefore, some studies have explored directly modeling incomplete data for end-to-end task performance.

\citeauthor{Che2018GRUD} proposed GRU-D, which directly models incomplete time series data by learning representations that account for missing patterns \cite{Che2018GRUD}.
\citeauthor{Hsu2020FuzzyCNN} developed FuzzyCNN for object detection in images with missing pixels \cite{Hsu2020FuzzyCNN}.
\citeauthor{habiba2020neurodesinfomissing} enhanced GRU-D by incorporating Neural ODEs to improve the modeling of incomplete data \cite{habiba2020neurodesinfomissing}. \citeauthor{Chang2023TN-ODE} proposed TN-ODE to handle irregular time intervals between sequential elements, facilitating the modeling of incomplete time series \cite{Chang2023TN-ODE}.
\citeauthor{Nasiri2022MFRFNN} employed a Multi-Functional Recurrent Fuzzy Neural Network (MFRFNN) to directly model complex, chaotic time series containing missing values \cite{Nasiri2022MFRFNN}.
\citeauthor{Zhang2023TriD-MAE} introduced TriD-MAE, which uses multi-period Temporal Convolutional Networks (TCNs) to capture correlations in incomplete time series data for direct future value prediction \cite{Zhang2023TriD-MAE}.
\citeauthor{Yu2024GinAR} designed GinAR, which utilizes an imputation-aware attention mechanism to learn spatio-temporal dependencies from incomplete data for subsequent forecasting tasks \cite{Yu2024GinAR}.

Despite these advances, representation learning methods specifically designed for incomplete multivariate spatio-temporal data remain an area requiring further exploration, particularly models that are robust to real-world data conditions such as variable missing rates.

\begin{figure*}
  \centering
  \includegraphics[width=\linewidth]{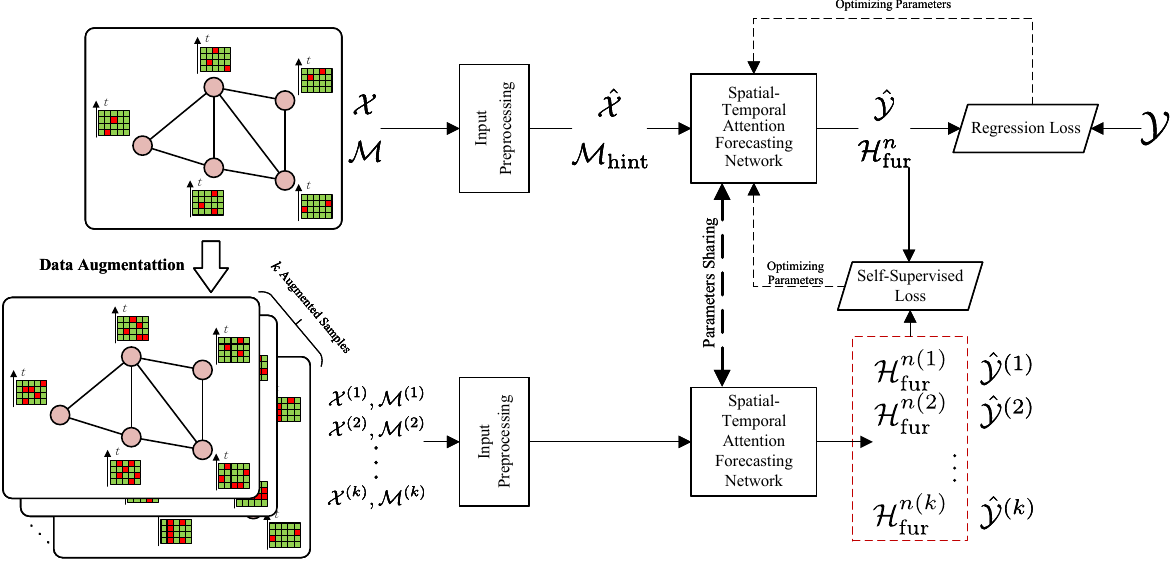}
  \caption{The framework of GeoMAE}
  \label{fig:attrmae:framework}
\end{figure*}

\section{Preliminaries}
\begin{table}
  \centering
  \caption{Frequently used notations}
  \label{tab:symb}
  \begin{adjustbox}{width=\linewidth}
    \begin{tabular}{cc}
      \toprule
      Notations                       & Description                                        \\
      \midrule
      $N_l$                           & The number of nodes in a spatio-temporal graph     \\
      $N_\mathrm{in}, N_\mathrm{out}$ & The number of historical inputs and future outputs \\
      $\mat{X}_t$                     & The reading matrix at timestamp $t$                \\
      $\mat{M}_t$                     & The missing indicator matrix at timestamp $t$      \\
      $\mathcal Y$                    & The output prediction                              \\
      \bottomrule
    \end{tabular}
  \end{adjustbox}
\end{table}
Table~\ref{tab:symb} shows the frequently used notations in this article. And we formulate our research problem as follows:
\begin{definition}[Spatio-Temporal Graph Prediction with Missing Values]
  Given continuous $N_{\mathrm{in}}$ readings of past moments $\mathcal X=[\mathbf X_1$, $\mathbf X_2$, $...$, $\mathbf X_{\mathrm{in}}]$, missing identification tensors $\mathcal M=[\mathbf M_1,\mathbf M_2,...,\mathbf M_{\mathrm{in}}]$ and static spacial relation $\mathcal G$,
  predicting the readings for the comming $N_{\mathrm{out}}$ steps, $\hat{\mathcal Y}=[\hat{\mathbf Y}_1,\hat{\mathbf Y}_2,...,\hat{\mathbf Y}_{\mathrm{out}}]$,
  among them, $\mathbf X_i\in\bbR^{N_l\times D_{\mathrm{in}}}$, $\mathbf M_i\in\{0,1\}^{N_l\times D_{\mathrm{in}}}$, $\mathbf Y_i\in\bbR^{N_l\times D_{\mathrm{out}}}$.
  Particularly, $\mathcal M$ represents whether there is a corresponding missing value in $\mathcal X$, 1 represents missing value while non-missing value is 0.
\end{definition}

\section{Methodologies}
GeoMAE consists of three major components: input data processing, spatio-temporal representation learning, and self-supervised auxiliary tasks.
Figure~\ref{fig:attrmae:framework} illustrates the primary workflow.
GeoMAE first uses a carefully designed input data preprocessing module to construct a hint tensor $\mathcal M_{hint}$ and performs necessary imputation on the input readings $\mathcal X$.
Then, it employs the \underline{s}patio-\underline{t}emporal \underline{a}ttention \underline{f}orecasting \underline{n}etwork (STAFN) for representation learning to obtain future spatio-temporal representation $\mathcal H_n$.
Finally, a fully connected network is used to decode $\mathcal H_n$ into the final prediction $\mathcal Y$.

Additionally, to enhance the stability and generalization ability of the spatio-temporal representation learning process, GeoMAE adopts a method similar to Masked AutoEncoder (MAE), i.e., adding extra masks to the data and incorporating an additional self-supervised loss to reduce the distance between the representations of the original data and the augmented samples.

The following subsections will introduce the input data preprocessing method, STAFN, and the multitask optimization method with self-supervised loss.

\subsection{Missing Data Preprocessing}
Unlike models based on decision trees that can handle missing values by treating them as special categorical features, deep learning models, particularly those based on neural networks, struggle to process input data with missing values, especially when missing values are represented by NaN (Not a Number).

Therefore, missing values require special preprocessing.
A common approach is to fill missing values with "0".
On the one hand, "0" is a passive signal in neural network computations, meaning that it does not actively activate or suppress neurons.
On the other hand, after the typical standardization preprocessing, the data would follow a standard normal distribution, and assigning zeros to missing values would have minimal impact on the original data distribution.
Moreover, adding extra missing values to the input data and filling them with 0s behaves similarly to Dropout in the input layer, which models uncertainty, helping the model cope with partial input loss \citep{Gal2016Dropout}.
However, this approach is only suitable when the missing data proportion is low and the actual missing rate in the data matches the training rate.
This is because a high missing rate would lead to a large number of zeros in the input samples, altering the data distribution, and this bias could be learned by the model, ultimately affecting its generalization ability.

Moreover, incorporating indicators of missingness in the input data is a feasible strategy.
GAIN inputs both the mask matrix and the data into the model to extract the data representation \citep{Yoon2018GAIN}, and BRITS constructs a missing matrix and inputs it into the model to better capture correlations in time series data \citep{Cao2018BRITS}.
Generally, these models construct a 0/1 mask matrix, where 0 represents missing data and 1 represents existing data (or vice versa).
However, this type of mask matrix faces a similar problem: the distribution of matrix values varies with different missing rates, which can affect the model's generalization ability.

Therefore, we propose a new data preprocessing method that includes two components: missing data imputation and mask matrix construction.
We first use random variables that follow a normal distribution for imputation to handle missing data.
\begin{equation}
  \hat{x}^{ijk}=\left\{
  \begin{array}{lr}
    x^{ijk}                              & m^{ijk}=0 \\
    \varepsilon\sim\mathcal N(0, \sigma) & m^{ijk}=1
  \end{array}
  \right.,
\end{equation}
where $i$, $j$, and $k$ are the indices of the input tensor $\mathcal X$ across the node, time, and feature dimensions, respectively. The variable $\varepsilon$ is the imputed random value that follows a normal distribution with a mean of 0 and a standard deviation of $\sigma$.

It is assumed that the input features have been standardized, i.e.,
\begin{equation}
  \mathbf x^{i}=\frac{\mathbf x^{i}_{\mathrm{raw}} - \bar{\mathbf{x}}^{i}_{\mathrm{raw}}}{\sigma(\mathbf x^{i}_{\mathrm{raw}})},
\end{equation}
where $\mathbf x^{i}_{\mathrm{raw}}$ represents the raw vector of the $i$th feature, $\bar{\mathbf{x}}^{i}_{\mathrm{raw}}$ represents the mean of that feature, and $\sigma(\mathbf x^{i}_{\mathrm{raw}})$ represents the standard deviation of that feature.
After this standardization process, the input data should follow a distribution with a mean of $0$ and a standard deviation of $1$.
Therefore, using random variables that follow a normal distribution with a mean of $0$ and a standard deviation of $\sigma$ for the imputation will minimally affect the distribution of the input features.
However, to avoid excessively large random values that could cause abnormal activation of certain neurons and introduce additional noise, it is recommended to choose a relatively small value for $\sigma$ (e.g., $0.2$).

Additionally, based on the identification tensor $\mathcal M$, this study constructed a HintSensor $\mathcal {M}_{\mathrm{hint}}$.
The specific process is as follows:
\begin{enumerate}
  \item Constructing a balanced mask tensor $\mathcal M_{\mathrm{sym}}=(m^{ijk}_{\mathrm{sym}})$, where
        \begin{equation}
          m^{ijk}_{\mathrm{sym}} = \left\{
          \begin{array}{lc}
            1  & m^{ijk} = 0 \\
            -1 & m^{ijk} = 1
          \end{array}
          \right.
        \end{equation}
  \item Standardize and scale $\mathcal M_{\mathrm{sym}}$ to obtain $\mathcal {M}_{\mathrm{hint}}$:
        \begin{equation}
          \mathcal {M}_{\mathrm{hint}} = \frac{\mathcal M_{\mathrm{sym}} - \bar{\mathcal M}_{\mathrm{sym}}}{\sigma\left[\mathcal M_{\mathrm{sym}}\right]}
        \end{equation}
        where $\bar{\mathcal M}_{\mathrm{sym}}$ and $\sigma\left[\mathcal M_{\mathrm{sym}}\right]$ Representing the mean and variance of $\mathcal M_{\mathrm{sym}}$ respectively.
\end{enumerate}

\subsection{Spatio-Temporal Attention Forecasting Network}
\label{sec:attrmae:stafn}

\begin{figure}[t]
  \centering
  \includegraphics[width=\linewidth]{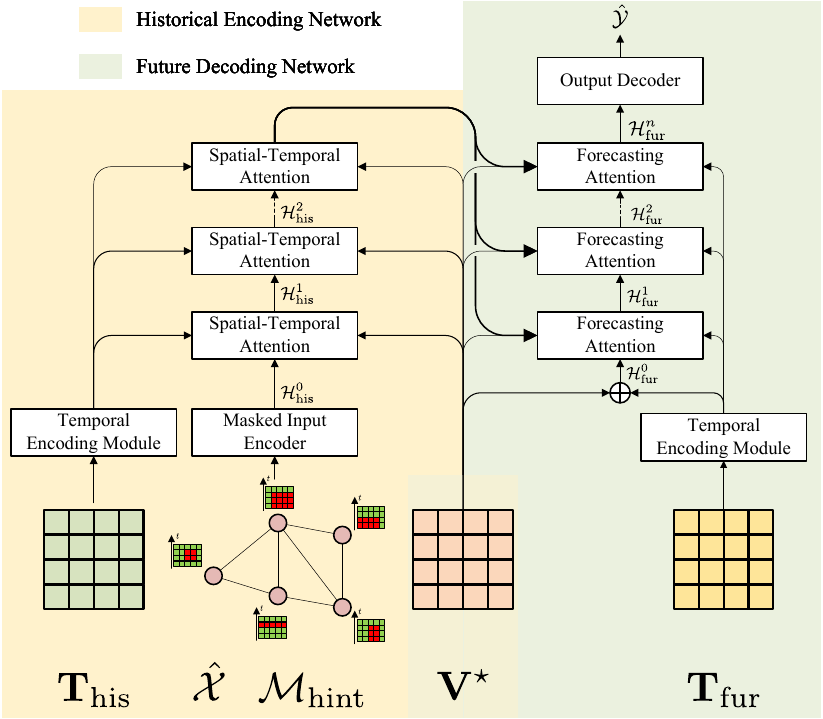}
  \caption{The structure of STAFN}
  \label{fig:attrmae:stafn}
\end{figure}

To better capture the complex and dynamic spatio-temporal correlations in the data, we design and implement a STAFN (\underline{S}patio-\underline{T}emporal \underline{A}ttention \underline{F}orecasting \underline{N}etwork) based on attention mechanisms. The main structure of STAFN is illustrated in Figure~\ref{fig:attrmae:stafn}.

STAFN adopts an Encoder-Decoder architecture, which comprises two parts: the historical encoding network (Encoder) and the future decoding network (Decoder). The temporal encoding module is shared between the encoder and the decoder. It aims to encode timestamp (e.g., month, day, and hour) into a temporal vector to help learn representation with multi-head attention mechanisms. The temporal encoding method is the common sin/cos positional encoding, as detailed in the literature \citep{Wu2023timesnet}.
Moreover, STAFN also maintains a learnable node representation denoted as $\mathbf V^{\star}\in\bbR^{N_l\times\Dmodel}$.
$\mathbf V^{\star}$ is fed into the spatio-temporal attention and forecasting attention modules, allowing attention mechanisms to capture spatial correlations from the data.
\begin{figure}
  \centering
  \null
  \hfill
  \subcaptionbox{Spatio-temporal attention module}[0.4\linewidth]{\includegraphics[scale=0.5]{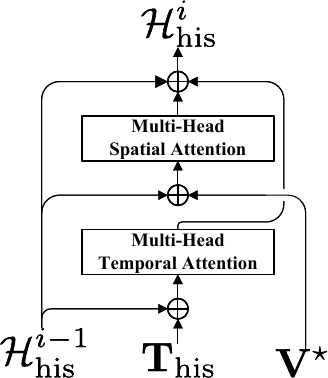}}
  \hfill
  \subcaptionbox{Forecasting attention module}[0.4\linewidth]{\includegraphics[scale=0.5]{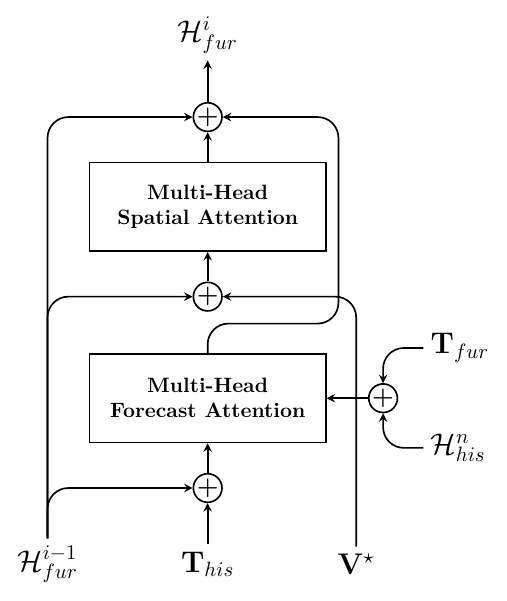}}
  \hfill
  \null
  \caption{The structures of two attention modules in STAFN}
  \label{fig:attrmae:attn-structures}
\end{figure}

STAFN consists of $n$ spatio-temporal attention modules and $n$ forecasting attention modules, where $n$ is a hyperparameter of the network.
These attention modules are composed of multi-head spatial attention, multi-head temporal attention, and multi-head forecast attention.
Figure~\ref{fig:attrmae:attn-structures} illustrates the structures of the spatio-temporal attention module and the forecasting attention module.

In the spatio-temporal attention module, the multi-head temporal attention mechanism and the multi-head spatial attention mechanism are arranged in parallel. The spatio-temporal representation $\mathcal H^{i-1}_{\mathrm{his}}$ output by the previous module is combined with the temporal vector $\mathbf T_{\mathrm{his}}$ and the spatial encoding $\mathbf V^{\star}$, and then fed into the multi-head attention mechanism.
For the forecasting attention module, since its initial representation $\mathcal H_{\mathrm{fur}}^{0}$ is merely the sum of the time encoding and the spatial encoding without any information related to the current sensor readings, a serial architecture is chosen, where the multi-head predictive attention mechanism is followed by the multi-head spatial attention mechanism.
This design ensures that the input to the spatial attention is not simply a combination of fixed spatial encodings and time encodings.

Here is a brief description of the multi-head spatial attention, temporal attention, and forecast attention:

\subsubsection{Multi-head Spatial Attention}

\begin{figure}[htbp]
  \centering
  \includegraphics[width=\linewidth]{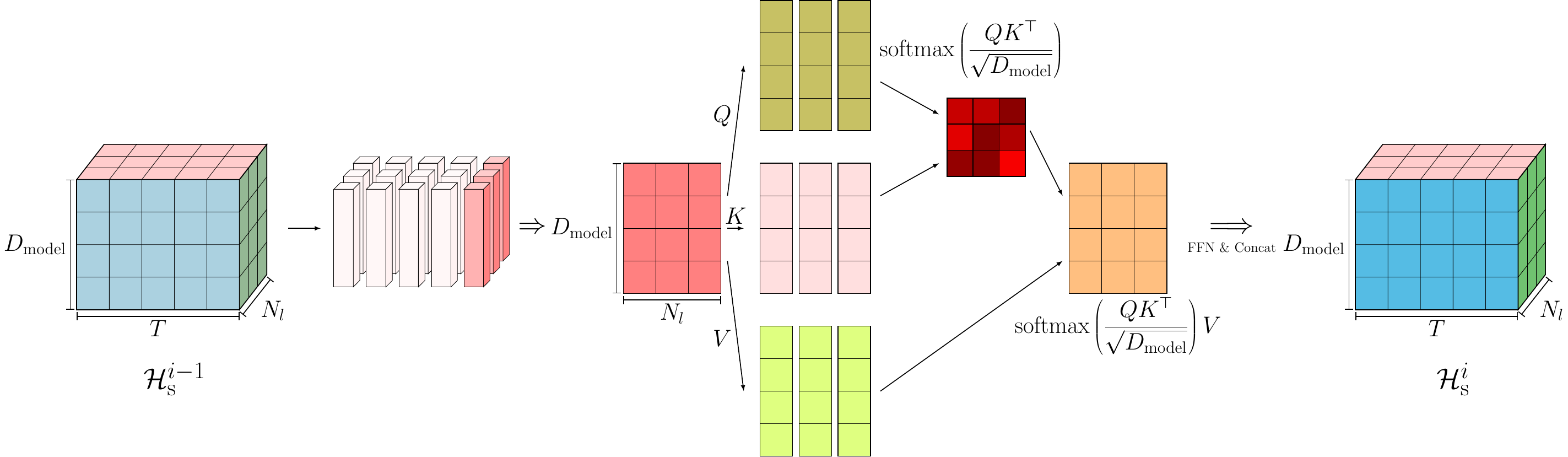}
  \caption{The structure of multi-head spatial attention}
  \label{fig:attrmae:spatial-attn}
\end{figure}

Figure~\ref{fig:attrmae:spatial-attn} illustrates the details of a spatial attention. $\mathcal H^{i-1}_{s}\in\bbR^{T\times N_l\times \Dmodel}$ is the input tensor to the spatial attention mechanism, containing time, space, and feature dimensions. We split $\mathcal H^{i-1}_{s}$ into $T$ representation matrices along the temporal dimension.
Without loss of generality, we take an arbitrary $\mathbf H_{s(j)}^{i-1}\in\bbR^{\Dmodel\times N_l}$ as an example to explain the computation of the spatial attention:
\begin{align}
  \mathbf Q_s^{i}       & =\mathbf W_{Q(s)}^{i}\mathbf H_{s(j)}^{i-1},                 \\
  \mathbf K_s^{i}       & =\mathbf W_{K(s)}^{i}\mathbf H_{s(j)}^{i-1},                 \\
  \mathbf V_s^{i}       & =\mathbf W_{V(s)}^{i}\mathbf H_{s(j)}^{i-1},                 \\
  \mathbf A_s^{i}       & = \frac{\mathbf Q_s^{i}\mathbf K_s^{i\top}}{\sqrt{\Dmodel}}, \\
  \alpha_{ef}           & = \frac{\exp{a_{ef}}}{\sum_{g=1}^{N_l}\exp{a_{eg}}},         \\
  \tilde{\mathbf A}     & = \left(\alpha_{ef}\right),                                  \\
  \mathcal H^{i}_{s(j)} & =\mathrm{MLP}(\tilde{\mathbf A}\mathbf V_s^{i}),
\end{align}
where, $\mathrm{MLP}(\cdot)$ denotes a multi-layer perceptron operation, and $\mathbf W_{Q(s)}^{i}$, $ \mathbf W_{K(s)}^{i}$, $\mathbf W_{V(s)}^{i}\in\bbR^{\Dmodel\times\Dmodel}$ represent the learnable parameters for the query (Q), key (K), and value (V) mappings, respectively.

\subsubsection{Multi-head Temporal Attention}
The temporal attention mechanism is similar to the spatial attention mechanism, except the spatial attention splits the tensor along the time axis and performs self-attention at the node dimension, while the temporal attention splits the representation along the node dimension and performs self-attention across different time steps for the same node. The computation process for a single attention head is as follows:
\begin{equation}
  \begin{array}{l}
    \mathcal H^{i}_{t(j)} = \mathrm{MLP}[
      \mathrm{softmax}\left(\frac{\mathbf W_{Q(t)}^{i}\mathbf H_{t(j)}^{i-1}\left(\mathbf W_{K(t)}^{i}\mathbf H_{t(j)}^{i-1}\right)^{\top}}{\sqrt{\Dmodel}}\right)\mathbf W_{V(t)}^{i}\mathbf H_{t(j)}^{i-1}],
  \end{array}
\end{equation}
Here, $\mathbf W_{Q(t)}^{i}$, $\mathbf W_{K(t)}^{i}$, $\mathbf W_{V(t)}^{i}\in\bbR^{\Dmodel\times\Dmodel}$ represent the learnable parameters for the Q, K, and V mappings, respectively.

\subsubsection{Multi-head Forecast Attention}
The forecasting attention is a special type of the temporal attention. Since future sensor readings are not unknown, directly applying the temporal attention mechanism to $\mathcal H_{fur}^{i-1}$ for Q, K, and V mappings is not reasonable. Therefore, we modify the self-attention mechanism as follows: First, the final output $\mathcal H_{\mathrm{his}}^{n}$ from the historical encoding network is used to calculate K and V, while $\mathcal H_{\mathrm{fur}}^{i-1}$ is used to calculate Q. Afterward, the attention computation is carried out using Q, K, and V. The formula is as follows:
\begin{equation}
  \begin{array}{l}
    \mathcal H^{i}_{fur(j)} = \mathrm{MLP}[
      \mathrm{softmax}\left(\frac{\mathbf W_{Q(t)}^{i}\mathbf H_{fur(j)}^{i-1}\left(\mathbf W_{K(t)}^{i}\mathbf H_{his(j)}^{n}\right)^{\top}}{\sqrt{\Dmodel}}\right)\mathbf W_{V(t)}^{i}\mathbf H_{his(j)}^{n}],
  \end{array}
\end{equation}
where, $\mathbf H_{his(j)}^{n}$ represents the spatio-temporal representation $\mathcal H_{\mathrm{his}}$ at node $j$.

\subsection{MAE Auxillary Task}
In incomplete spatio-temporal graph prediction, the data distribution may change over time. The missing rates and patterns all pose challenges to spatio-temporal representation learning.
They may severely impact the model's generalization performance.

To minimize the negative effects of these factors on the model's generalization performance, we adopt a self-supervised auxiliary task based on random masking for spatio-temporal graph representation learning with missing data:
$k$ augmented samples $(\mathcal X^{(1)}, \mathcal M^{(1)})$, $(\mathcal X^{(2)}, \mathcal M^{(2)})$, ..., $(\mathcal X^{(k)}, \mathcal M^{(k)})$ are generated from $\mathcal X$ and $\mathcal M$ by adding extra masks to simulate different missing patterns and rates. Then, the above $k$ augmented samples are fed into STAFN to obtain the corresponding spatio-temporal representations $\mathcal H_{fur}^{n(1)}$, $\mathcal H_{fur}^{n(2)}$, ..., $\mathcal H_{fur}^{n(k)}$. Finally, an L2 loss function is used to reduce the distance between $\mathcal H_{fur}^{n(1)}$, $\mathcal H_{fur}^{n(2)}$, ..., $\mathcal H_{fur}^{n(k)}$ and the spatio-temporal representation of the raw sample $\mathcal H_{fur}^{n}$.

In particular, the mutual approach of $\mathcal H_{fur}^{n(i)}$ and $\mathcal H_{fur}^{n}$ is controlled by a parameter $\varphi$.
The additional self-supervised loss, after being scaled by the hyperparameter $\lambda$, is added to the final optimization in the form of multi-task learning, i.e.,
\begin{align}
  \mathcal L_{\mathrm{reg}} & = \lVert\hat{\mathcal Y} - \mathcal Y \rVert,                                                                                                                                               \\
  \mathcal L_{\mathrm{MAE}} & = \frac{1}{k}\sum_{i=1}^{k}\lVert\mathcal H_{fur}^{n(i)} - \left[\mathcal H_{fur}^{n}\right] \rVert_2 + \varphi\lVert \mathcal H_{fur}^{n} - \left[\mathcal H_{fur}^{n(i)}\right] \rVert_2, \\
  \mathcal L_{tot}          & = \mathcal L_{reg} + \lambda\cdot\mathcal L_{\mathrm{MAE}},
\end{align}
where $\lVert\cdot\rVert$ represents the regression loss function, which can be L1, L2, or other loss functions depending on the task. In particular, $\lVert\cdot\rVert_2$ represents the L2 loss function. Additionally, $\left[\cdot\right]$ represents the constant operation, that is, the gradient of this item is not calculated during the gradient backpropagation.

\section{Evaluations}
We validated GeoMAE on a real-world dataset.
Section~\ref{sec:attrmae:exp:settings} introduces the dataset, task settings, and model hyperparameter settings.
Section~\ref{sec:attrmae:exp:overall-performance} presents a comparison of the main performance indicators between GeoMAE and other baseline models.
Section~\ref{sec:attrmae:exp:abla} conducts ablation experiments on multiple modules of GeoMAE.

\subsection{Experimental Settings}
\label{sec:attrmae:exp:settings}
\subsubsection{Dataset and Task Settings}

We use a real-world dataset to validate our proposed model, and the dataset is described as follows:
\begin{description}[leftmargin=6pt]
  \item[BJ-Air] It includes 35 air quality monitoring stations in Beijing, 6 air pollutants, and 6 meteorological conditions. The data spans from 2015-01-01 to 2017-12-31, with a sampling interval of 1 hour, 26,304 timestamps in total. Since the concentration of PM2.5 varies greatly and often determines AQI, it is used as the target for prediction.
\end{description}

The experiments in this section split the BJ-Air dataset by year, i.e., data from 2015, 2016, 2017 is served as the training, validation, and test set, respectively.
In this way, they all fully cover the four seasons of the year, allowing for a more balanced test of the model's performance.
Moreover, the number of input and output time steps is set to 12.

\begin{figure}
  \centering
  \includegraphics[width=\linewidth]{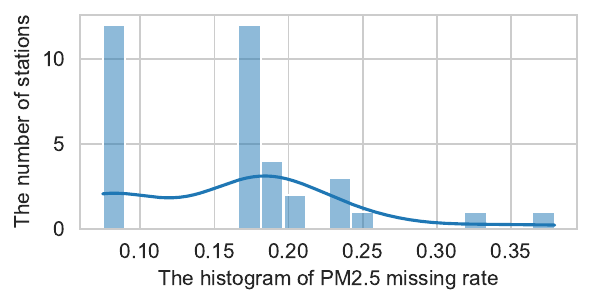}
  \caption{The distribution of missing rates of 35 Beijing air monitoring stations}
  \label{fig:attrmae:beijing-air-missing-rates}
\end{figure}

Specifically, since we focus on incomplete data representation learning, the following settings are made for the missing values in the data:
\begin{description}[leftmargin=6pt]
  \item[Missing Rate]
        Considering that the BJ-Air data already has a certain amount of missing values (Figure~\ref{fig:attrmae:beijing-air-missing-rates} shows the distribution of missing rates in this dataset), the fixed missing rates for the BJ-Air dataset are set at 25\%, 50\%, 75\%, and 90\%.
        Specifically, the above missing rates are set only for the input sample $\mathcal X$, and the model is required to predict the complete $\mathcal Y$.
  \item[Missing Pattern]
        For the training set, this study establishes a combined missing scenario incorporating point-wise, row-wise, and column-wise missing data. For the validation and test sets, two types of missing scenarios are designed: point missing and block missing. The block sizes for the latter include $4\times 4$ and $8\times 8$, representing missing data from 4 sensor readings over 4 consecutive time steps and 8 sensor readings over 8 consecutive time steps, respectively. In the block missing scenario, regions and block sizes are randomly selected to mask the data. This process repeats until the overall missing rate of the data reaches the specified requirement (e.g., 30\%, 40\%).
\end{description}

\subsubsection{Baseline Models}
After reviewing recent work on missing value imputation methods and spatiotemporal prediction models with missing data, this study selected and implemented several well-established algorithms from different categories as baseline models. The specific methods are introduced as follows:
\begin{itemize}
  \item \textbf{MLP}~~
        A fully connected neural network that flattens all input time steps, nodes, and different types of readings into a single vector, then uses multiple fully connected layers to learn data representations and output predictions. MLP is the most fundamental representation learning method, theoretically capable of approximating any function and capturing spatiotemporal correlations without bias.
  \item \textbf{LSTM}~~
        Utilizes an Encoder-Decoder architecture. The inputs from different nodes and attributes are flattened into a vector, which is then processed by multiple LSTM layers to learn representations and output predictions. The LSTM model based on the Encoder-Decoder design is one of the most classic neural network architectures in time series processing. It captures temporal dependencies via LSTM units. Considering correlations between sensor readings at different spatial locations, the LSTM model uses a fully connected network for preliminary encoding of sequences from all locations.
  \item \textbf{GRU-D}\cite{Che2018GRUD}~~
        GRU-D (Gated Recurrent Unit with Decay) is a variant of the GRU model designed for handling irregularly sampled time series data with missing values. Since the original GRU-D network does not consider multi-layer stacking or decoding future sequences, the same Encoder-Decoder architecture as the LSTM model is used here. The first layer of the Encoder employs GRU-D units, subsequent Encoder layers use standard GRU units, and the Decoder is entirely composed of GRU units.
  \item \textbf{AGCRN}\cite{Bai2020AGCRN}~~
        AGCRN employs a dynamic graph structure learning module and Node Adaptive Parameter Learning Graph Convolutional Networks (NAPL-GCN), fully considering the dynamic, complex, and uncertain relationships between traffic nodes in the real world. It is a classic and representative excellent algorithm for spatiotemporal graph prediction tasks.
  \item \textbf{GWNet}\cite{wu2019graphwavenet}~~
        Graph WaveNet uses adaptive graph convolution and temporal convolution networks to capture spatial and temporal dependencies in spatiotemporal graph data, respectively. By simply adjusting hyperparameters, it can achieve highly competitive performance on many spatiotemporal graph prediction tasks.
  \item \textbf{TriD-MAE}\cite{Zhang2023TriD-MAE}~~
        TriD-MAE is a general-purpose multivariate time series pre-training model that effectively handles multivariate time series forecasting with missing values using TCN modules.
  \item \textbf{STAFN}~~
        A model constructed using only the backbone network STAFN from GeoMAE, excluding GeoMAE's input preprocessing and self-supervised auxiliary tasks. STAFN adopts the popular Transformer-based architecture \cite{vaswani2017attention, Zhou2021Informer, Liang2023AirFormer}, capturing temporal and spatial correlations through self-attention networks.
\end{itemize}

\subsubsection{Implementation}
We implement GeoMAE with 4 spatio-temporal attention modules, 4 forecasting attention modules, $\Dmodel=512$, 4 augmented samples per sample. The weight parameter $\varphi$ is 0.25, and $\lambda$ is 0.75. An AdamW optimizer with a learning rate of 0.0002 and a regularization strength of 0.001 is used. Moreover, an NVIDIA RTX4090 24GiB GPU was used for acceleration.

\subsubsection{Evaluation Metrics}
we employ three common evaluation metrics in regression tasks to compare performance, i.e., MAE, RMSE, and SMAPE, and they can be fomulated as
\begin{equation}
  \mathrm{MAE}=\frac{1}{n}\sum_{i=1}^{n}\lvert\hat y^i - y^i\rvert.
  \label{eq:mae}
\end{equation}
\begin{equation}
  \mathrm{RMSE}=\sqrt{\frac{1}{n}\sum_{i=1}^{n}(\hat y^i - y^i)^2}.
  \label{eq:rmse}
\end{equation}
\begin{equation}
  \mathrm{SMAPE} = \frac{100\%}{n} \sum_{t=1}^{n} \frac{|\hat y^i - y^i|}{|\hat y^i|+|y^i|},
  \label{eq:smape}
\end{equation}

\subsection{Performance Comparison}
\label{sec:attrmae:exp:overall-performance}
\begin{figure}[htbp]
  \centering
  \null
  \hfill
  \subcaptionbox{MLP}[.32\linewidth]{\includegraphics[width=\linewidth]{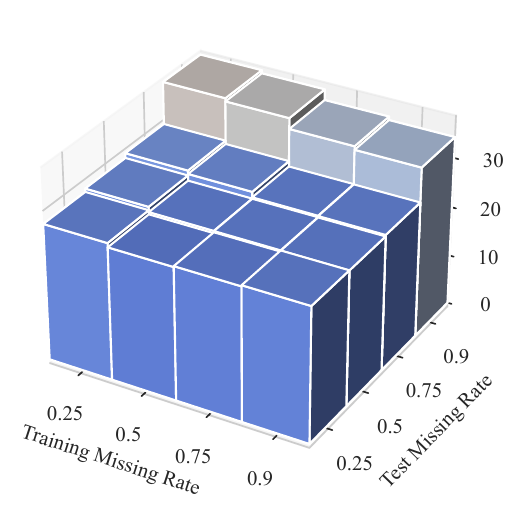}}
  \hfill
  \subcaptionbox{LSTM}[.32\linewidth]{\includegraphics[width=\linewidth]{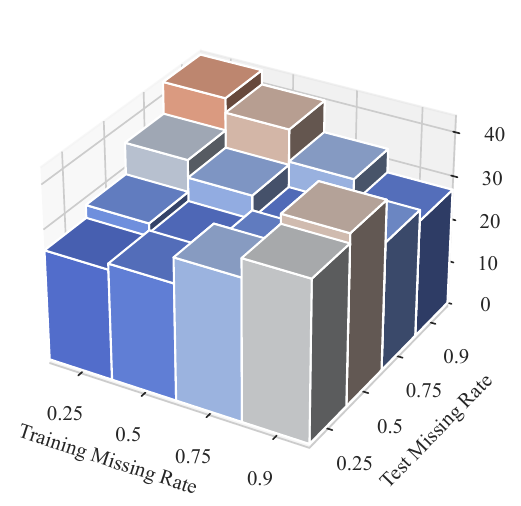}}
  \hfill
  \subcaptionbox{STAFN}[.32\linewidth]{\includegraphics[width=\linewidth]{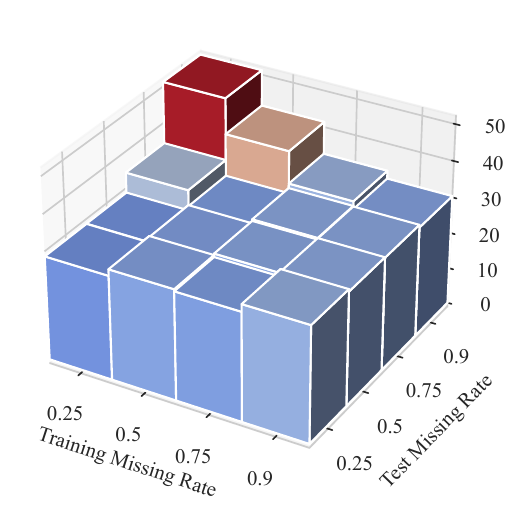}}
  \hfill
  \null
  \caption{The performance of MLP, LSTM, and STAFN under different training-testing missing propotions.}
  \label{fig:attrmae:3dplots}
\end{figure}

\begin{figure}[htbp]
  \centering
  \null
  \hfill
  \subcaptionbox{MLP}[0.32\linewidth]{\includegraphics[width=\linewidth]{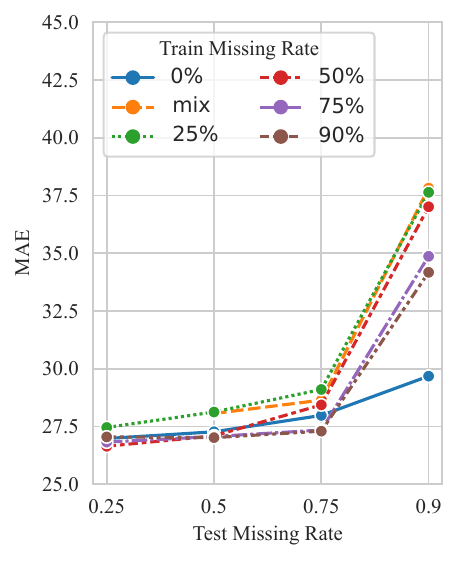}}
  \hfill
  \subcaptionbox{LSTM}[0.32\linewidth]{\includegraphics[width=\linewidth]{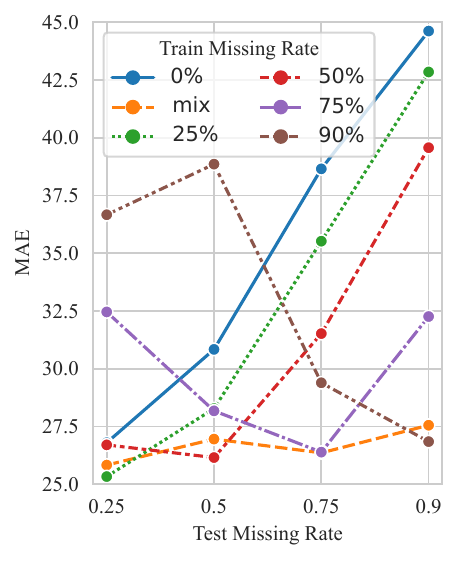}}
  \hfill
  \subcaptionbox{STAFN}[0.32\linewidth]{\includegraphics[width=\linewidth]{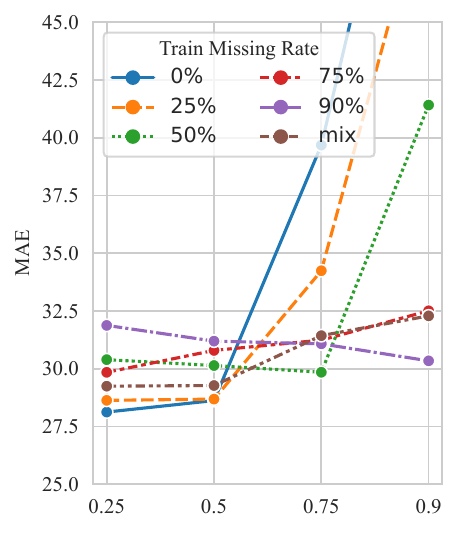}}
  \hfill
  \null
  \caption{The curves of performance for MLP, LSTM, and STAFN.}
  \label{fig:attrmae:mixed-masking-rate}
\end{figure}

First, this study investigates the impact of different training and test missing rates on model performance under the point-wise missing scenario. Figure~\ref{fig:attrmae:3dplots} illustrates the performance of three models under various missing rate configurations. For a more objective discussion of the models' performance and characteristics, the experimental results presented here are the averages from five repeated runs with fixed random seeds.
From Figure~\ref{fig:attrmae:3dplots}(a), it can be observed that the MLP model demonstrates relatively stable performance across most scenarios, with MAE remaining between 27 and 29, except for the case with a 90\%test missing rate. It shows a good ability to handle distribution shifts between training and test data, indicating robust generalization performance. In contrast, the LSTM-based Encoder-Decoder model exhibits a greater reliance on the consistency of data distributions. It achieves excellent performance when the test data missing rate matches the training missing rate, with MAE hovering around 25~27 (refer to Figure~\ref{fig:attrmae:mixed-masking-rate}), significantly outperforming the MLP model under the same conditions. However, its performance degrades noticeably when the missing rates are mismatched.
Although STAFN employs the more flexible self-attention mechanism, the common data splitting method in spatio-temporal forecasting tasks easily leads to distribution inconsistencies between the training, validation, and test sets. This makes models based on self-attention prone to overfitting to noise patterns present only in the training set, consequently impairing their generalization ability. This viewpoint, initially raised in \cite{liang2022basicts}, is corroborated by the experimental results of this study. As shown in Figure~\ref{fig:attrmae:3dplots}(c), STAFN's performance on the test set is poor, even inferior to the simple MLP model, indicating significant overfitting. Furthermore, the STAFN model also demonstrates a dependency on data distribution consistency.

Given that both the LSTM and STAFN models exhibit a significant dependency on the consistency of missing ratios between the training and test sets, a more feasible approach is to avoid fixing the missing ratio in the training set, thereby allowing the training data to simulate as many missing ratios as possible.
Figure~\ref{fig:attrmae:mixed-masking-rate} displays the performance curves under different training set configurations. The curve labeled "Mixed" represents the results obtained without a fixed missing rate in the training set. It can be observed that increasing the diversity of the training data effectively enhances the model's capability to handle data with different missing rates, significantly improving its generalization ability.
However, the cost of achieving this more "stable" performance is a slight compromise in peak performance, which is somewhat inferior to the results achieved when the test set missing rate matches the training set missing rate.

\begin{table}[htbp]
  \centering
  \caption{Performance comparison on point missing datasets}
  \label{tab:attrmae:overall-performance}
  \begin{adjustbox}{width=\linewidth}
    \begin{tabular}{cccccc}
      \toprule
      \multirow{2}[0]{*}{Model}    & \multirow{2}[0]{*}{Metrics} & \multicolumn{4}{c}{Missing Ratio}                                                                                     \\
                                   &                             & 25\%                              & 50\%                      & 75\%                      & 90\%                      \\
      \midrule
      \multirow{3}[0]{*}{MLP}      & MAE                         & 26.98$\pm$0.07                    & 27.27$\pm$0.11            & 27.98$\pm$0.16            & 29.68$\pm$0.15            \\
                                   & RMSE                        & 46.07$\pm$0.72                    & 46.46$\pm$0.89            & 46.63$\pm$0.6             & 49.03$\pm$0.06            \\
                                   & SMAPE                       & 0.35$\pm$0.00                     & 0.35$\pm$0.00             & 0.36$\pm$0.00             & 0.38$\pm$0.00             \\
      \midrule
      \multirow{3}[0]{*}{LSTM}     & MAE                         & 25.83$\pm$0.81                    & 26.95$\pm$0.21            & 26.36$\pm$0.28            & 27.55$\pm$0.87            \\
                                   & RMSE                        & 43.68$\pm$1.10                    & 44.46$\pm$0.47            & 44.02$\pm$0.78            & 45.43$\pm$1.58            \\
                                   & SMAPE                       & 0.33$\pm$0.02                     & 0.35$\pm$0.00             & 0.34$\pm$0.01             & 0.36$\pm$0.01             \\
      \midrule
      \multirow{3}[0]{*}{GRU-D}    & MAE                         & 25.94$\pm$0.99                    & 25.68$\pm$0.74            & 26.86$\pm$1.29            & 29.08$\pm$0.88            \\
                                   & RMSE                        & 42.61$\pm$1.33                    & 42.61$\pm$0.96            & 44.48$\pm$1.59            & 46.59$\pm$1.20            \\
                                   & SMAPE                       & 0.35$\pm$0.01                     & 0.34$\pm$0.02             & 0.36$\pm$0.02             & 0.38$\pm$0.01             \\
      \midrule
      \multirow{3}[0]{*}{AGCRN}    & MAE                         & 27.17$\pm$0.44                    & 26.86$\pm$0.54            & 26.5$\pm$0.29             & 27.5$\pm$0.72             \\
                                   & RMSE                        & 45.88$\pm$0.61                    & 45.81$\pm$0.7             & 44.76$\pm$0.67            & 45.5$\pm$0.78             \\
                                   & SMAPE                       & 0.35$\pm$0.01                     & 0.34$\pm$0.01             & 0.34$\pm$0.01             & 0.35$\pm$0.01             \\
      \midrule
      \multirow{3}[0]{*}{GWNet}    & MAE                         & 25.01 $\pm$ 0.23                  & 24.15 $\pm$ 0.01          & 24.66 $\pm$ 0.02          & 29.44 $\pm$ 0.22          \\
                                   & RMSE                        & 42.83 $\pm$ 0.39                  & 41.67 $\pm$ 0.24          & 41.87 $\pm$ 0.37          & 45.75 $\pm$ 0.31          \\
                                   & SMAPE                       & 0.33 $\pm$ 0.00                   & 0.32 $\pm$ 0.00           & 0.34 $\pm$ 0.00           & 0.41 $\pm$ 0.00           \\
      \midrule
      \multirow{3}[0]{*}{TriD-MAE} & MAE                         & 26.18 $\pm$ 0.39                  & 26.08 $\pm$ 0.66          & 27.76 $\pm$ 0.06          & 31.50 $\pm$ 0.05          \\
                                   & RMSE                        & 45.12 $\pm$ 0.09                  & 45.02 $\pm$ 0.24          & 47.23 $\pm$ 0.02          & 51.29 $\pm$ 0.07          \\
                                   & SMAPE                       & 0.35 $\pm$ 0.00                   & 0.35 $\pm$ 0.00           & 0.36 $\pm$ 0.00           & 0.40 $\pm$ 0.00           \\
      \midrule
      \multirow{3}[0]{*}{GeoMAE}   & MAE                         & \textbf{22.42 $\pm$ 0.01}         & \textbf{22.35 $\pm$ 0.12} & \textbf{23.57 $\pm$ 0.17} & \textbf{26.13 $\pm$ 0.35} \\
                                   & RMSE                        & \textbf{40.01 $\pm$ 0.57}         & \textbf{40.39 $\pm$ 0.74} & \textbf{41.56 $\pm$ 0.08} & \textbf{43.35 $\pm$ 0.14} \\
                                   & SMAPE                       & \textbf{0.30 $\pm$ 0.00}          & \textbf{0.30 $\pm$ 0.00}  & \textbf{0.32 $\pm$ 0.00}  & \textbf{0.35 $\pm$ 0.00}  \\
      \bottomrule
    \end{tabular}%
  \end{adjustbox}
\end{table}%

When the training missing rate mismatches the test missing rate, the performance of most baseline models degrades significantly. 
Therefore, employing a variable missing rate in the training set is an effective strategy to mitigate this issue.

Table~\ref{tab:attrmae:overall-performance} compares the performance of the baselines and GeoMAE when trained with a variable missing rate. 
The results show that GRU-D, the only baseline algorithm specifically designed for irregular time series data with missing values, performs poorly on this task. 
This is likely because the input data is equally spaced, rendering its capability to handle irregular time intervals ineffective. 
AGCRN, which combines RNNs and GCNs with an adaptive graph mechanism, demonstrates strong fitting capabilities. However, it also suffers from severe overfitting due to the distribution shift between the training and test sets. 
In contrast, GWNet, another representative spatio-temporal graph model, achieves competitive performance with better generalization ability. 
TriD-MAE, which also utilizes a masked autoencoding mechanism, shows mediocre performance. 
A potential reason is its lack of dedicated structures for capturing spatial correlations, such as graph neural networks.

Finally, augmented by its self-supervised auxiliary loss and input data preprocessing mechanisms, the complete GeoMAE framework achieves the best performance across datasets with different missing rates. 
This outcome demonstrates its effectiveness in learning spatio-temporal representations directly from incomplete data for downstream tasks. 
Furthermore, it underscores the significant contribution of the self-supervision and preprocessing modules in enhancing the generalization capability of the representation learning model.

\begin{table}[htbp]
  \centering
  \caption{Performance comparison on block missing datasets}
  \label{tab:attrmae:block-missing}
  \begin{adjustbox}{width=\linewidth}
    \begin{tabular}{cccccc}
      \toprule
      \multirow{2}[0]{*}{Model}    & \multirow{2}[0]{*}{Metrics} & \multicolumn{4}{c}{Missing Ratios}                                                                                     \\
                                   &                             & 30\%                               & 40\%                      & 50\%                      & 60\%                      \\
      \midrule
      \multirow{3}[0]{*}{MLP}      & MAE                         & 25.02 $\pm$ 0.12                   & 25.90 $\pm$ 0.46          & 26.26 $\pm$ 0.36          & 27.02 $\pm$ 0.54          \\
                                   & RMSE                        & 43.64 $\pm$ 0.45                   & 44.48 $\pm$ 0.60          & 45.10 $\pm$ 0.16          & 46.38 $\pm$ 0.60          \\
                                   & SMAPE                       & 0.32 $\pm$ 0.00                    & 0.33 $\pm$ 0.00           & 0.33 $\pm$ 0.00           & 0.34 $\pm$ 0.00           \\
      \midrule
      \multirow{3}[0]{*}{LSTM}     & MAE                         & 25.70 $\pm$ 0.08                   & 25.80 $\pm$ 0.10          & 25.93 $\pm$ 0.23          & 26.69 $\pm$ 0.13          \\
                                   & RMSE                        & 43.94 $\pm$ 0.27                   & 44.08 $\pm$ 0.34          & 44.47 $\pm$ 1.03          & 44.99 $\pm$ 0.40          \\
                                   & SMAPE                       & 0.33 $\pm$ 0.00                    & 0.33 $\pm$ 0.00           & 0.33 $\pm$ 0.00           & 0.34 $\pm$ 0.00           \\
      \midrule
      \multirow{3}[0]{*}{AGCRN}    & MAE                         & 25.94 $\pm$ 0.03                   & 26.03 $\pm$ 0.13          & 26.46 $\pm$ 0.15          & 26.71 $\pm$ 0.14          \\
                                   & RMSE                        & 44.60 $\pm$ 0.08                   & 45.07 $\pm$ 0.63          & 45.28 $\pm$ 0.07          & 45.87 $\pm$ 0.54          \\
                                   & SMAPE                       & 0.33 $\pm$ 0.00                    & 0.33 $\pm$ 0.00           & 0.33 $\pm$ 0.00           & 0.33 $\pm$ 0.00           \\
      \midrule
      \multirow{3}[0]{*}{GRU-D}    & MAE                         & 23.51 $\pm$ 0.01                   & 23.74 $\pm$ 0.06          & 23.83 $\pm$ 0.05          & 24.22 $\pm$ 0.14          \\
                                   & RMSE                        & 40.43 $\pm$ 0.57                   & 40.41 $\pm$ 0.44          & 40.74 $\pm$ 0.27          & 40.42 $\pm$ 0.01          \\
                                   & SMAPE                       & 0.32 $\pm$ 0.00                    & 0.33 $\pm$ 0.00           & 0.33 $\pm$ 0.00           & 0.33 $\pm$ 0.00           \\
      \midrule
      \multirow{3}[0]{*}{GWNet}    & MAE                         & 23.74 $\pm$ 0.13                   & 23.69 $\pm$ 0.02          & 23.96 $\pm$ 0.19          & 24.48 $\pm$ 0.02          \\
                                   & RMSE                        & 41.62 $\pm$ 0.75                   & 41.91 $\pm$ 0.20          & 41.73 $\pm$ 0.80          & 42.31 $\pm$ 0.17          \\
                                   & SMAPE                       & 0.32 $\pm$ 0.00                    & 0.32 $\pm$ 0.00           & 0.32 $\pm$ 0.00           & 0.33 $\pm$ 0.00           \\
      \midrule
      \multirow{3}[0]{*}{TriD-MAE} & MAE                         & 25.41 $\pm$ 0.13                   & 25.49 $\pm$ 0.01          & 25.95 $\pm$ 0.02          & 27.06 $\pm$ 0.04          \\
                                   & RMSE                        & 44.25 $\pm$ 0.20                   & 44.55 $\pm$ 0.14          & 44.95 $\pm$ 0.06          & 45.72 $\pm$ 0.01          \\
                                   & SMAPE                       & 0.34 $\pm$ 0.00                    & 0.34 $\pm$ 0.00           & 0.34 $\pm$ 0.00           & 0.36 $\pm$ 0.00           \\
      \midrule
      \multirow{3}[0]{*}{GeoMAE}   & MAE                         & \textbf{22.76 $\pm$ 0.04}          & \textbf{22.50 $\pm$ 0.20} & \textbf{22.53 $\pm$ 0.04} & \textbf{22.88 $\pm$ 0.08} \\
                                   & RMSE                        & \textbf{39.78 $\pm$ 0.17}          & \textbf{40.31 $\pm$ 1.06} & \textbf{40.26 $\pm$ 0.11} & \textbf{40.82 $\pm$ 0.25} \\
                                   & SMAPE                       & \textbf{0.31 $\pm$ 0.00}           & \textbf{0.30 $\pm$ 0.00}  & \textbf{0.31 $\pm$ 0.00}  & \textbf{0.31 $\pm$ 0.00}  \\
      \bottomrule
    \end{tabular}
  \end{adjustbox}
\end{table}%

Furthermore, Table~\ref{tab:attrmae:block-missing} presents the performance of GeoMAE and the baseline models on the block missing datasets. Notably, since the original dataset has a missing rate of approximately 20\% and it might be challenging to generate valid datasets when the target block missing rate is set too high, the missing rates for the block missing study are configured between 30\% and 60\%.
According to Table~\ref{tab:attrmae:block-missing}, the forecasting difficulty under the block missing scenario is slightly lower than that on point missing datasets with equivalent overall missing rates. 
This phenomenon can likely be attributed to the data generation process for block missingness. As it is not possible to directly generate a mask matrix that precisely meets the target missing rate, the data generation process for block missing involves iterative steps until the actual missing rate of the data satisfies the requirement. This results in the actual missing rate in point missing datasets being slightly higher than that in block missing datasets, thereby increasing the task difficulty. 
Consequently, the performance of all baseline models improves compared to their results on the point missing datasets.
Notably, GRU-D performs significantly better on the block missing datasets. This is possibly because its mechanism for handling irregular time intervals assists in better managing the information loss characteristic of block missing data, leading to improved forecasting performance. As for the GeoMAE model, its performance remains relatively stable and shows no significant fluctuations across the 30\% to 60\%missing rate range. 
It is important to highlight that none of the training data included corresponding block missing scenarios. 
Therefore, GeoMAE can effectively handle complex and varying block missing scenarios, learning robust and reliable spatio-temporal representations directly from incomplete data.
\subsection{Ablation Studies}
\label{sec:attrmae:exp:abla}
This section further verifies the impact of the input preprocessing module and input random masking strategy on model performance through ablation experiments, with the specific variant models compared as follows:
\begin{itemize}
  \item \textbf{\{LSTM,AGCRN,GRU-D\}+MAE}~Replacing STAFN with \{ LSTM, AGCRN, GRU-D \}.
  \item \textbf{GeoMAE-FM}~Fixing training set missing ratio to 50\%.
  \item \textbf{GeoMAE-NM}~Using "0" imputing without mask tensor.
  \item \textbf{GeoMAE-01}~Using "0" imputing and "01" masking.
\end{itemize}

\begin{figure}
  \centering
  \subcaptionbox{Impacts of different backbone networks}[0.48\linewidth]{\includegraphics[width=\linewidth]{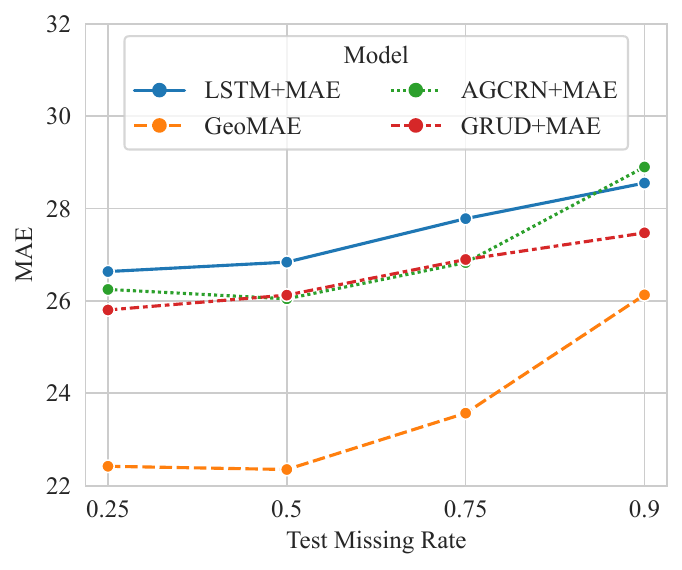}}
  \subcaptionbox{GeoMAE with different masking methods}[0.48\linewidth]{\includegraphics[width=\linewidth]{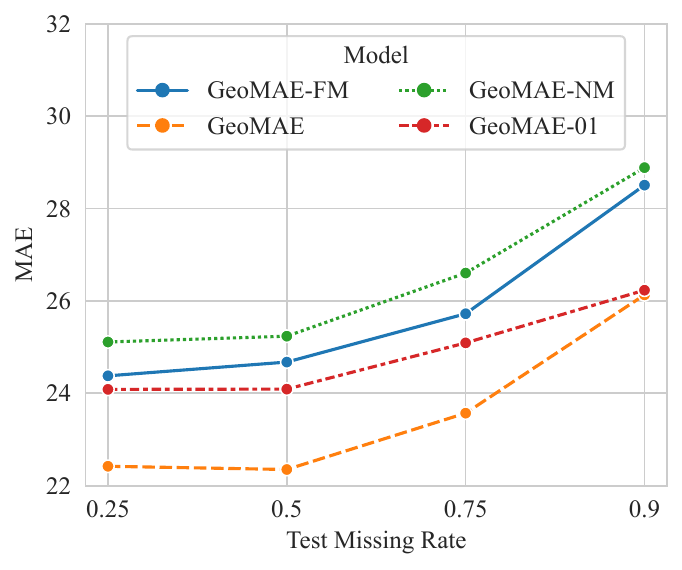}}
  \caption{The curves of performance for GeoMAE and its variants}
  \label{fig:attrmae:abla}
\end{figure}

Figure~\ref{fig:attrmae:abla}(a) shows the performance curves of four different backbone networks after adding the input preprocessing mechanism and self-supervised loss.
Compared with the performance in Table~\ref{tab:attrmae:overall-performance}, less effective improvement is achieved in other scenarios except for AGCRN performing better at low missing ratios.
This indicates that LSTM, AGCRN, and GRU-D have limited capabilities when used as representation learning networks.
STAFN should be the most suitable backbone network for this task.

Secondly, from Figure~\ref{fig:attrmae:abla}(b), it can be seen that there is a significant performance gap between GeoMAE-NM and GeoMAE.
This demonstrates that even with the self-supervised auxiliary loss, missing values significantly affect GeoMAE-NM, highlighting the importance of the proposed preprocessing method.
The introduction of the 01 masking makes the performance of the GeoMAE-01 model noticeably better than models without any masking mechanism.
On this basis, the newly proposed incomplete data preprocessing method helps GeoMAE better mitigate the impact of missing values on representation learning, further enhancing its accuracy in predicting future readings.
On the other hand, the GeoMAE-FM model trained with a fixed missing rate shows a significant performance gap compared to GeoMAE.
This indicates that even with additional self-supervised loss, a fixed missing rate can still introduce certain biases.
Once the model encounters a missing rate that does not match the training set in practical applications, the model's prediction accuracy will decline significantly.
The above experimental results show that the random missing rate and the prompt matrix construction method in GeoMAE can effectively enhance the model's generalization ability, thereby improving its performance in practical implementation.

\subsection{Hyperparameter Sensitivity}
\begin{figure}[htbp]
    \centering
    \subcaptionbox{The number of augmented samples' effect}[0.31\linewidth]{\includegraphics[width=\linewidth]{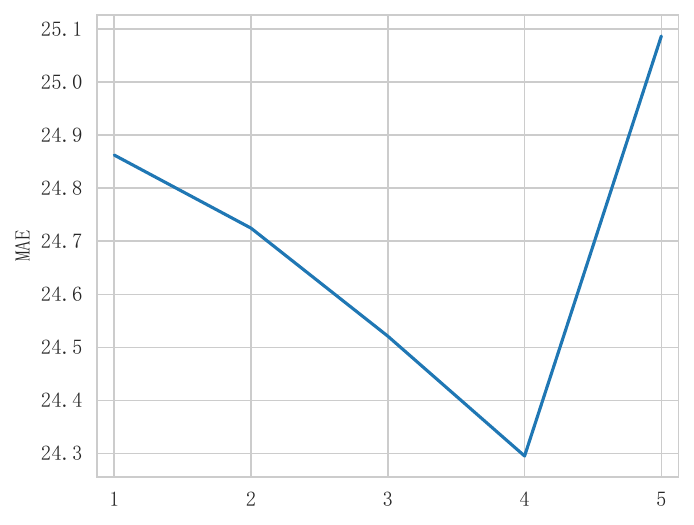}}
    \subcaptionbox{The effect of $D_{\mathrm{model}}$}[0.31\linewidth]{\includegraphics[width=\linewidth]{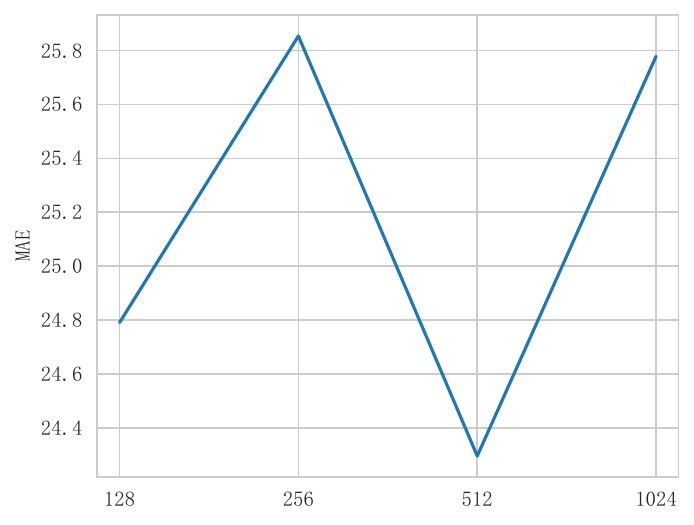}}
    \subcaptionbox{The effect of batch sizes}[0.31\linewidth]{\includegraphics[width=\linewidth]{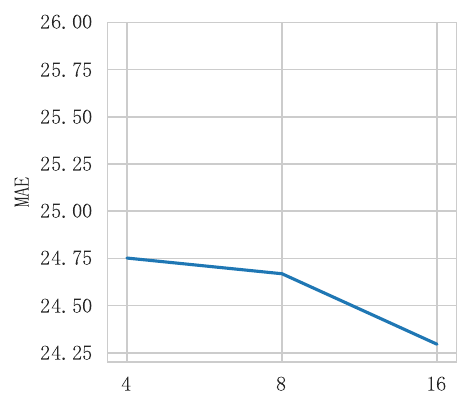}}
    \caption{The effects of hyperparameters}
    \label{fig:attrmae:hp-test}
\end{figure}

In this study, we conducted hyperparameter sensitivity tests under a condition where 75\% of the test set data were missing. The results are shown in Figure~\ref{fig:attrmae:hp-test}, with the y-axis representing the Mean Absolute Error (MAE) as the evaluation metric.

As illustrated in Figure~\ref{fig:attrmae:hp-test}(a), the accuracy of the GeoMAE model gradually improves with an increase in the number of augmented samples ($1\rightarrow 4$). However, when the number of augmented samples increases to 5, there is a decline in model performance. A plausible explanation for this phenomenon could be attributed to the use of a smaller batch size during training due to memory constraints, which compromises the stability of the training process and hinders the model from converging to its optimal state. Therefore, taking into account the overall performance, four augmented samples emerge as the optimal choice for the BJ-Air dataset.

The outcomes related to model size, depicted in Figure~\ref{fig:attrmae:hp-test}(b), do not exhibit a clear trend. Specifically, the accuracy at a model size of 256 is slightly lower than that at a size of 128. This observation might be attributed to normal random variation. With standard deviations of MAE across five experiments with different random seeds being 0.40 and 0.37 for model sizes of 128 and 256 respectively, these values exceed the average MAE difference between them. Hence, it can be inferred that there is no significant disparity in prediction accuracy between model sizes of 128 and 256. However, a notable improvement in accuracy is observed when the model size escalates to 512, which is likely associated with increased model capacity. Interestingly, when the model size further increases to 1024, the prediction error rises instead, suggesting potential overfitting. Another possible reason for this may be similar to the scenario with five augmented samples, where a reduced batch size due to limited memory affects model convergence. Consequently, considering model capacity and computational cost, a model size of 512 appears to be more suitable for this task.

Lastly, given the preceding findings on the potential impact of batch size on model convergence, we further explored how varying batch sizes affect model robustness, as shown in Figure~\ref{fig:attrmae:hp-test}(c). The results indicate that the performance of the GeoMAE model improves with an increase in batch size. Due to memory limitations, we did not investigate whether larger batch sizes could further enhance model performance. The influence of batch size on model robustness warrants further investigation in future studies.

\section{Conclusion}
In this article, we focus on the problem of modeling spatio-temporal graph data with missing values, proposing a new representation learning model GeoMAE to capture the complex and dynamic spatio-temporal correlations.
It attempts to address the challenges of the changes in missing ratios and patterns caused by maintenance conditions in practical applications to representation learning by designing new input data preprocessing methods and mask-based self-supervised auxiliary tasks.
We validate GeoMAE on a real-world dataset.
The experimental results show that the GeoMAE model exhibits consistent and stable performance in complex missing scenarios.
Its self-supervised auxiliary loss effectively enhances the robustness and generalization ability of the learned spatio-temporal representations.

In the future, we will conduct experiments on more datasets to validate the generalization ability of GeoMAE.

\bibliographystyle{apalike}
\bibliography{references}

\end{document}